\documentclass[letterpaper, 10 pt, conference]{ieeeconf}
\IEEEoverridecommandlockouts
\usepackage{cite}
\usepackage{amsmath,amssymb,amsfonts}
\usepackage{algorithmic}
\usepackage{graphicx}
\usepackage{textcomp}
\usepackage{xcolor}

\usepackage{microtype}
\usepackage{subcaption}
\usepackage{booktabs} 
\usepackage{hyperref}
\usepackage{url}

\usepackage{mathtools}

\usepackage{textcomp}
\usepackage{float}
\usepackage{xspace}
\usepackage{bm}
\usepackage{multirow}
\usepackage{lipsum}
\usepackage{mathtools}
\usepackage{cuted}
\usepackage{colortbl}
\usepackage[frozencache,cachedir=.]{minted}
\usepackage{tcolorbox}
\usepackage{censor}

\definecolor{Green}{RGB}{102,252,102}

\def\BibTeX{{\rm B\kern-.05em{\sc i\kern-.025em b}\kern-.08em
    T\kern-.1667em\lower.7ex\hbox{E}\kern-.125emX}}

\title{Compositional Average Velocity Modeling for Efficient Diffusion Offline Reinforcement Learning}

\author{Thanh Nguyen$^{1}$ and Abu Hanif Muhammad Syarubany$^{1}$ and Hwanhee Kim$^{1}$ and Chang D. Yoo $^{1}$
\thanks{*This work was supported by the Institute for Information \& Communications Technology Planning \& Evaluation (IITP) grant funded by the Korea government (MSIT) (No. RS-2021-II211381) and the National Research Foundation of Korea (NRF) grant funded by MSIT (No. RS-2025-24742969).}
\thanks{$^{1}$ The authors are with the School of Electrical Engineering, KAIST, Daejeon, Republic of Korea.
        {\tt\small }}%
}

\begin{document}

\maketitle
\thispagestyle{empty}
\pagestyle{empty}

\begin{abstract}
Diffusion-based policies have recently emerged as a powerful paradigm for offline reinforcement learning, but their reliance on multi-step denoising introduces substantial computational overhead and requires backpropagation through time during training, limiting scalability and practical deployment. In this work, we propose \textbf{Instant Flow Q-Learning (InsQL)}, a single-step generative policy learning framework that retains the expressive capacity of diffusion models while eliminating iterative sampling and BPTT. InsQL builds on the flow-matching perspective of diffusion policies, but replaces marginal velocity parameterization with \emph{average-velocity} modeling and introduces a principled compositional mechanism, yielding a theoretically grounded objective that learns average velocity consistently across scales. This formulation provides a direct noise-to-action mapping and enables single-step action generation. InsQL further unifies compositional average-velocity modeling with behavior-regularized actor--critic optimization in a single framework, requiring only a single policy network and avoiding distillation, multi-stage procedures, and Jacobian computations during training. At inference time, actions are generated in a single forward pass and can be refined via value-guided candidate selection to improve performance. Extensive experiments on standard offline RL benchmarks and real-world robotic manipulation tasks show that InsQL matches or surpasses state-of-the-art diffusion-based methods while substantially reducing both training and inference time. Overall, these results establish compositional average-velocity modeling as an efficient and scalable alternative to multi-step generative policies for offline RL.

\end{abstract}

\section{Introduction}
Offline reinforcement learning (RL) aims to learn decision-making policies from fixed datasets, enabling deployment in domains where online exploration is unsafe, costly, or infeasible, including robotics, autonomous driving, and healthcare \cite{levine2020offline}. A growing focus in offline RL is the development of policies that are both \emph{expressive} and \emph{efficient}, especially in complex control settings where behavior data can induce multimodal action distributions \cite{wang2023diffusion,park2025flow}. Generative policy classes—most notably diffusion-based models—have recently emerged as a compelling alternative to Gaussian actors by transforming a simple base distribution into a flexible action distribution through learnable transformations \cite{dong2024cleandiffuser,fujimoto2021minimalist,kumar2020conservative,kostrikov2021offline,nguyen2026fast}. This increased representational capacity has translated into strong empirical improvements over Gaussian policies on widely used offline RL benchmarks \cite{dong2024cleandiffuser,wang2023diffusion}.

Despite these gains, most diffusion-style policies remain tightly coupled to \emph{iterative generation}, producing actions through multi-step refinement from noise \cite{wang2023diffusion}. This dependency introduces substantial computational overhead during training, where action sampling is repeatedly required throughout optimization \cite{kang2023efficient,wang2023diffusion}. Moreover, when combined with value-based objectives, multi-step generation can impose additional optimization burdens—often involving backpropagation through time (BPTT) across solver steps—which may increase sensitivity to hyperparameters and affect final performance \cite{park2025flow}. At deployment, the same multi-step procedure translates into non-trivial inference latency, posing a challenge for real-time closed-loop control. Existing work has attempted to mitigate these issues through faster samplers \cite{kang2023efficient}, alternative learning paradigms (e.g., IQL-style updates) \cite{hansen2023idql}, and acceleration mechanisms such as auxiliary policies or distillation-based training \cite{park2025flow,chen2023score,chen2024diffusion,lu2025habitizing}, sometimes involving Jacobian-related computations \cite{nguyen2026onestep}. While effective in certain regimes, these approaches often introduce extra design choices, training dependencies, or multi-stage pipelines that complicate scaling and impose trade-offs among simplicity, efficiency, and policy quality.

We propose \textbf{Instant Flow Q-Learning (InsQL)}, an end-to-end offline RL framework that enables \emph{single-step} action generation during both training and inference. InsQL builds on the flow-matching interpretation of diffusion policies, but is explicitly designed to avoid iterative denoising and solver unrolling. Instead of learning an instantaneous marginal velocity field at intermediate timesteps, InsQL parameterizes the policy using the \emph{average velocity over finite intervals}, allowing the full noise-to-action transformation to be queried directly. To make this representation consistent and learnable across interval lengths, we introduce a \emph{velocity composition} mechanism that recursively composes short-interval average velocities into a coherent long-range update; in the infinitesimal limit, this structure recovers the marginal velocity field learned by standard flow matching. As a result, when integrated into a behavior-regularized actor--critic objective, InsQL eliminates solver unrolling and backpropagation through time (BPTT) over denoising trajectories, yielding a simpler and substantially more compute-efficient learning procedure while preserving expressive capacity. Finally, to better exploit value information at test time, we incorporate a lightweight \emph{Q-guided sampling} strategy that improves action quality at a modest additional cost. Empirically, InsQL achieves state-of-the-art or competitive performance on standard D4RL benchmarks while significantly reducing training time and increasing decision frequency. We further demonstrate its practicality in real-world robotic manipulation, where InsQL attains higher success rates than prior methods.

In summary, our main contributions are:
\begin{itemize}
    \item \textbf{Compositional average-velocity policy modeling.}
    We introduce a novel policy parameterization based on \emph{average velocity over finite intervals}, together with a principled velocity composition mechanism that recursively constructs long-range noise-to-action mappings while remaining compatible with Q-learning.

    \item \textbf{A unified and efficient one-step actor--critic framework.}
    We develop an end-to-end offline RL framework that enables single-step action generation during both training and inference, without BPTT, solver unrolling, multi-stage pipelines, explicit distillation, or Jacobian-based optimization. The framework further incorporates lightweight Q-guided sampling at test time to improve action quality with modest latency overhead.

    \item \textbf{Strong empirical performance and efficiency.}
    InsQL achieves competitive or superior performance on D4RL benchmarks while significantly reducing training time and increasing decision frequency, and it further demonstrates improved success rates in real-world robotic manipulation.
\end{itemize}

\section{Related Work}

Offline reinforcement learning (Offline RL) aims to learn decision-making policies from static datasets without additional environment interaction \cite{levine2020offline,nguyen2024towards,nguyen2024uncertainty,luu2024mitigating,luu2022visual}. The paradigm has recently been explored in conjunction with online reinforcement learning
\cite{schulman2017proximal,nguyen2021sample,nguyen2021robust} to offer a more sample-efficient learning paradigm. To mitigate distributional shift and overestimation, early value-based offline RL methods introduced conservative objectives, including CQL \cite{kumar2020conservative}, TD3+BC \cite{fujimoto2021minimalist}, and IQL \cite{kostrikov2021offline}. While often effective, these approaches typically rely on unimodal Gaussian actors, which can struggle to represent the multimodal action distributions induced by heterogeneous behavior data.

Motivated by their ability to model complex, high-dimensional distributions, diffusion models \cite{ho2020denoising,song2021scorebased} have recently been incorporated into offline RL in several ways. Prior work has used diffusion models for trajectory synthesis and planning \cite{janner2022diffuser,ajay2022conditional}, as expressive policy parameterizations \cite{wang2023diffusion,hansen2023idql}, and as generative augmentation tools \cite{zhu2023diffusion}. Among policy-based methods, Diffusion Q-Learning (DQL) \cite{wang2023diffusion} is a prominent baseline that replaces the Gaussian actor in TD3+BC with a diffusion policy to better capture multimodal action spaces. Follow-up studies such as Clean Diffuser \cite{dong2024cleandiffuser} and recent large-scale evaluations \cite{lu2025makes,lu2025habitizing} report that DQL remains consistently competitive relative to both planner-style diffusion approaches and conventional policy-learning baselines. However, DQL's reliance on iterative denoising introduces non-trivial computational overhead during both training and inference \cite{kang2023efficient}, and its interaction with value-based optimization can lead to sensitivity and occasionally suboptimal convergence \cite{park2025flow}.

Several lines of work attempt to address these limitations. One direction accelerates sampling by designing improved solvers that reduce the number of denoising steps required by diffusion policies \cite{kang2023efficient}. Another avoids backpropagation through time by training a diffusion policy primarily via behavior cloning and using a separately trained IQL-style value function to reweight or select actions \cite{hansen2023idql}. A complementary strategy distills a multi-step diffusion policy into a single-step actor \cite{chen2023score,chen2024diffusion}. However, IQL-style learning can underperform actor--critic alternatives in some regimes \cite{park2024value}. To obtain single-step inference while retaining actor--critic updates, \cite{park2025flow} replaces diffusion with a flow-based policy trained to imitate behavior and then distills it into a one-step actor; however, the distillation stage still requires repeated queries to the underlying iterative generator. In addition, recent one-step formulations based on MeanFlow \cite{nguyen2026onestep,wang2025one,geng2025mean} typically involve explicit Jacobian-related computations, which have been argued to suffer from explosion or poor conditioning in high dimensions \cite{holderrieth2024hamiltonian}.

In contrast, our goal is to remove the iterative bottleneck directly by training an end-to-end \emph{single-step} generative policy that preserves expressiveness while avoiding multi-stage training procedures, explicit distillation, and Jacobian computation.

\section{Preliminaries}
\begin{figure*}[htb]
\centering
\includegraphics[width=1.0\linewidth]{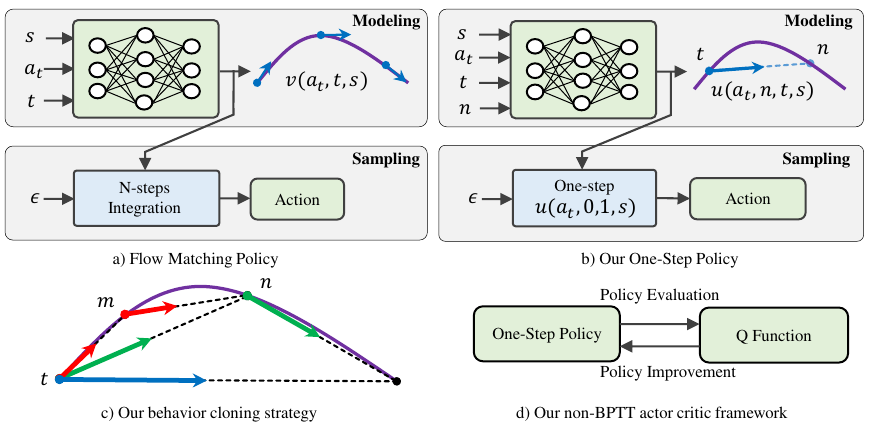}
    \caption{\textbf{Instant Flow Q-Learning (InsQL) overview.} 
\textbf{(a)} Standard flow-matching (diffusion) policies learn an instantaneous marginal velocity field $v(a_t,t,s)$ and generate actions by multi-step integration. 
\textbf{(b)} InsQL instead parameterizes the policy with an \emph{average-velocity} field $u(a_t,n,t,s)$ over finite intervals, enabling \emph{single-step} action generation by directly mapping noise to action. 
\textbf{(c)} \emph{Velocity composition}: long-interval average velocity (blue) is recursively composed from shorter-interval average velocities (green/red); in the infinitesimal limit, this recovers the marginal velocity used in FM. 
\textbf{(d)} InsQL unifies the one-step policy into a non-BPTT behavior-regularized actor--critic framework, with optional Q-guided candidate selection for test-time refinement.}
\vspace{-0.5cm}
\label{fig:framework}
\end{figure*}

\textbf{Offline Reinforcement Learning.}
Reinforcement learning (RL) is typically formalized as a Markov decision process (MDP) $\mathcal{M}=(\mathcal{S},\mathcal{A},P,R,\gamma)$, where $\mathcal{S}$ and $\mathcal{A}$ are the state and action spaces, $P(s'\!\mid s,a)$ denotes environment dynamics, $R(s,a)$ is the reward, and $\gamma\in[0,1)$ discounts future returns. RL aims to learn a parameterized policy $\pi_\theta(a\mid s)$ that maximizes the expected discounted cumulative reward
\begin{equation}
\mathbb{E}_{\pi_\theta}\!\left[\sum_{h=0}^{\infty}\gamma^{h}R(s_h,a_h)\right].
\end{equation}
For supporting policy optimization, the action--value function is often introduced:
\begin{equation}
Q^\pi(s,a)=\mathbb{E}_{\pi}\!\left[\sum_{h=0}^{\infty}\gamma^{h}R(s_h,a_h)\,\middle|\, s_0=s,\ a_0=a\right],
\end{equation}
which evaluates the long-term return of executing $a$ at $s$ and then following $\pi$.

\emph{Offline} RL is a branch of RL where the agent has no access to online interaction and must learn entirely from a fixed batch of transitions
$\mathcal{D}=\{(s_h,a_h,s_{h+1},r_h)\}$ collected by unknown behavior policies. A central challenge is \emph{multimodality}: because offline datasets are often collected under diverse (and suboptimal) strategies, the induced action distribution can be highly multimodal, making it difficult for unimodal policy parameterizations to capture. Effective offline RL therefore requires improving the policy while modeling the diverse behaviors present in the dataset.

\textbf{Behavior-regularized actor--critic.}
A standard and practically strong approach is behavior-regularized actor--critic, which iterates between (i) \emph{policy evaluation} via a critic that estimates $Q$, and (ii) \emph{policy improvement} via an actor that maximizes critic values under an explicit constraint toward the dataset distribution.
Concretely, the critic is trained with a double-Q Bellman regression objective:
\begin{equation}
\small
\resizebox{0.95\linewidth}{!}{$
 \mathcal{L}_Q(\phi)
 = \mathbb{E}_{(s,a,r,s')\sim \mathcal{D},\ a'\sim \pi_{\theta}}\!\left[
 \left(r+\gamma \min_{i} Q_{\phi_i'} (s',a')-Q_{\phi_i} (s,a)\right)^2
 \right],$}
\end{equation}
where $\phi'$ are slowly-updated target parameters for stability. The actor is updated to prefer high-value actions while remaining close to the behavior distribution:
\begin{equation}
\resizebox{0.85\linewidth}{!}{$
\mathcal{L}(\theta)
= -\mathbb{E}_{(s,a)\sim\mathcal{D},\ a^\pi\sim\pi_\theta}\!\left[
\alpha\,Q_{\phi}(s,a^\pi) + \log \pi_\theta(a^\pi\mid s)
\right]
$}
\end{equation}
where $\alpha$ balances value maximization and behavior regularization through a log-likelihood term.

While this framework is conceptually simple, its computational cost is dominated by \emph{how} the policy generates actions during both actor and critic updates. In particular, diffusion-style policies produce actions through multiple iterative refinement steps; training then requires repeated multi-step sampling, and gradients from the actor loss must backpropagate through the entire generation trajectory, leading to expensive backpropagation-through-time (BPTT). This makes both policy improvement and evaluation significantly slower.

\textbf{Modeling Policies with Flow Matching.}
The generation of the DQL \cite{wang2023diffusion} is based on SDE, which is hard to improve, we can employ deterministic generative policies built on flow matching (FM), which learns an ODE that transports noise to data.
Given data $(s,a)\sim\mathcal{D}$ and noise $\epsilon\sim\mathcal{N}(\mathbf{0},\boldsymbol{I})$, a common FM construction uses a linear path:
\begin{equation}
a_t=(1-t)a+t\epsilon,\qquad
v_{\mathrm{cnd}}=\epsilon-a,\qquad t\in[0,1],
\end{equation}
and trains a neural velocity field $v_\theta(a_t,t;s)$, as illustrated in Fig.~\ref{fig:framework} (a), by minimizing the conditional FM objective:
\begin{equation}
\mathbb{E}_{t\sim U[0,1],\ (s,a)\sim\mathcal{D},\ \epsilon}\left\|
v_\theta(a_t,t;s)-v_{\mathrm{cnd}}(a_t\mid a,\epsilon;s)
\right\|^2.
\end{equation}
At test time, actions are generated by integrating the learned ODE $\frac{d a_t}{d t} = v_\theta(a_t, t; s)$ from $a_1=\epsilon$ to $a_0$ (e.g., using Euler updates $a_{t-\Delta}=a_t-\Delta\,v_\theta(a_t,t;s)$).

A tempting idea is to set $\Delta=1$ and obtain a \emph{single-step} policy. However, the learned marginal velocity field typically induces globally curved transport, so a one-shot Euler step can be inaccurate even if the multi-step ODE solution is correct. Consequently, existing flow-based policies still often rely on multiple integration steps, and when embedded in actor--critic training they continue to incur BPTT-like costs through the solver.


\section{Methodology}

We now introduce \textbf{Instant Flow Q-Learning (InsQL)}, designed to satisfy three requirements: high expressive capacity for multimodal action distributions, single-step action generation for efficient training and inference without BPTT, and a simple framework without distillation, multi-stage training, or Jacobian computation.

Unlike conventional flow-matching approaches that learn marginal velocity fields at infinitesimal time scales, InsQL parameterizes the policy in terms of the average velocity over finite time intervals. As illustrated in Fig.~\ref{fig:framework}(a,b), the marginal velocity represents the instantaneous tangent direction of a trajectory, whereas our formulation captures the finite-interval average velocity between two time points.

A key structural property of average velocity is compositionality: the average velocity over a long interval can be constructed by composing average velocities over shorter sub-intervals. As shown in Fig.~\ref{fig:framework}(c), a long-range average velocity (blue) can be decomposed into intermediate average velocities (green), each of which can in turn be decomposed into shorter ones (red). This hierarchical decomposition yields a well-defined limiting behavior: as the interval length approaches zero, the average velocity converges to the marginal velocity field. Standard flow matching therefore emerges naturally as the infinitesimal base case, providing both a principled boundary condition and a natural termination criterion for the recursive construction.

Leveraging this structure, InsQL uses a single shared network to model average velocity across all temporal scales. Training is anchored at the marginal-velocity level to learn short-interval average velocities, which are then recursively composed to represent longer intervals. Through this aggregation, the model ultimately learns the full noise-to-action average velocity, enabling direct single-step action generation while preserving the expressive power of flow-based generative policies.



\textbf{Policy Modeling.} Mathematically,
we consider a continuous-time generative process defined over $t \in [0,1]$, where $a_t \in \mathbb{R}^d$ denotes the action variable at time $t$. The process interpolates between the clean action at $t=0$ and pure noise at $t=1$ for a given state $s$, and is governed by a marginal velocity field $v(a_t, t; s)$ satisfying $\frac{d a_t}{d t} = v(a_t, t; s).$ Instead of directly parameterizing the marginal velocity field, we model the \emph{average velocity} over a finite interval $[n,t]$, defined as:
\begin{equation}
u(a_t, n, t; s)
\;\triangleq\;
\frac{1}{t-n}
\int_{n}^{t} v(a_\tau, \tau; s)\, d\tau,
\qquad 0 \le n < t \le 1.
\end{equation}
The average velocity represents the normalized net displacement induced by the flow between times $n$ and $t$, and serves as the fundamental quantity in our formulation.

\paragraph{Compositional Consistency}
A key structural property of $u$ follows directly from the additivity of definite integrals. 
For any $0 \le n \le m \le t \le 1$, we have
\begin{equation}
\begin{aligned}
u(a_t, n, t; s)
&= \frac{1}{t-n}
\int_n^t v(a_\tau, \tau; s)\, d\tau \\
&= \frac{1}{t-n}
\left(
\int_n^m v(a_\tau, \tau; s)\, d\tau
+
\int_m^t v(a_\tau, \tau; s)\, d\tau
\right) \\
&=
\frac{m-n}{t-n} u(a_m, n, m; s)
+
\frac{t-m}{t-n} u(a_t, m, t; s).
\end{aligned}
\end{equation}

Equivalently,
\begin{equation}
u(a_t, n, t; s)
=
\beta\, u(a_m, n, m; s)
+
(1-\beta)\, u(a_t, m, t; s),
\end{equation}
where $\beta \triangleq \frac{m-n}{t-n} \in [0,1]$.

This relation establishes a compositional consistency: the average velocity over a long interval is a convex combination of average velocities over sub-intervals. This property enables recursive construction of long-range dynamics from shorter ones.

\paragraph{Boundary Conditions}
To anchor learning to the true local dynamics of the flow and avoid degenerate solutions, the average velocity must recover the marginal velocity in the infinitesimal limit. Specifically,
\begin{equation}
\lim_{n \to t} u(a_t, n, t; s)
=
v(a_t, t; s).
\label{eq:boundary_condition}
\end{equation}

Thus, when the interval length shrinks to zero, the average velocity coincides with the marginal velocity field. This provides a principled boundary condition and connects our formulation to standard FM.

In practice, the boundary condition in Eq.~\eqref{eq:boundary_condition} requires accurate estimates of the marginal velocity. Rather than relying on a pretrained teacher model, we exploit the conditional velocity $v_{\mathrm{cnd}}$ naturally derived from FM. This yields a fully self-contained training objective in which short-interval average velocities are anchored directly to the conditional velocity, eliminating the need for external supervision.

\paragraph{Final Behavior Cloning Objective} Let $u_\theta$ denote a neural parameterization of the average velocity.
Given a sampled transition $(s,a)\sim\mathcal D$, noise $\epsilon\sim\mathcal N(0,I)$, and time $t\sim U[0,1]$, we construct $a_t = (1-t)a + t\epsilon$. For times $0 \le n \le m \le t \le 1$, calculate $\beta = \frac{m-n}{t-n}$. We then define compositional loss:
\begin{equation}
a_m
=
a_t - (t-m)\,u_\theta(a_t,m,t;s),
\label{eq:am_def}
\end{equation}
\begin{equation}
\tilde u_\theta
=
\beta\, u_\theta(a_m,n,m;s)
+
(1-\beta)\, u_\theta(a_t,m,t;s).
\label{eq:utilde_def}
\end{equation}
\begin{equation}
\mathcal{L}_{\mathrm{cps}}(\theta)
=
\mathbb{E}_{\mathcal D,\,\epsilon,\,t,\,n,\,m}
\left[
\left\|
u_\theta(a_t,n,t;s)
-
\operatorname{sg}(\tilde u_\theta)
\right\|_2^2
\right],
\label{eq:lcomp}
\end{equation}
where $\operatorname{sg}(\cdot)$ denotes the stop-gradient operator.
The stop-gradient prevents recursive gradient propagation through the composed target while enforcing compositional consistency.

The boundary condition can be incorporated as a regularization term using a small $\Delta$ for stability:
\begin{equation}
\resizebox{\columnwidth}{!}{$
\mathcal{L}_{\mathrm{bdr}}(\theta)
=
\mathbb{E}_{\mathcal D,\,\epsilon,\,t}
\left[
\left\|
u_\theta(a_t,t-\Delta,t;s)
-
\operatorname{sg}\!\left(
v_{\mathrm{cnd}}(a_t,t \mid a,\epsilon;s)
\right)
\right\|_2^2
\right]
$}
\label{eq:bnd}
\end{equation}
Finally, the overall behavior cloning objective is defined as
\begin{equation}
\mathcal L_{BC}(\theta)
=
(1-\lambda)\,\mathcal L_{\mathrm{cps}}
+
\lambda\,\mathcal L_{\mathrm{bdr}},
\label{Eq:BC}
\end{equation}
where $\lambda \in [0,1]$ controls the probability of enforcing the boundary constraint.

\paragraph{Single-Step Policy Sampling.}

The policy $\pi_\theta$ is implicitly defined through the learned full-interval average velocity.
Given noise $\epsilon \sim \mathcal N(0,I)$, a one-step action sample is generated as
\begin{equation}
\pi_\theta(s)
\;\triangleq\;
\epsilon
-
u_\theta(\epsilon,0,1;s).
\end{equation}
Under the true dynamics, the average velocity over $[0,1]$ equals the displacement $\epsilon - a$.
Therefore, $\epsilon - u(\epsilon,0,1;s) = a,$
which recovers the clean action in a single step.

\textbf{Instant Flow Q-Learning.}
We construct InsQL by integrating the proposed one-step policy parameterization and behavior cloning objective into a behavior-regularized actor–critic framework, as illustrated in Fig.~\ref{fig:framework} (d). The behavior regularization term is given by $\mathcal{L}_{\mathrm{BC}}(\theta)$, as defined in Eq.~\ref{Eq:BC}.


The critic and actor are optimized using the following objectives:
\begin{equation}
\small
 \mathcal{L}_Q(\phi) = \mathbb{E}_{ \mathcal{D}, {a}' \sim \pi_{\theta}} \left[ \left( r + \gamma \min_{i} Q_{\phi_i'}({s}', {a}') - Q_{\phi_i}({s}, {a}) \right)^2 \right],
 \label{eq:bfq_critic_loss}
\end{equation}
\begin{equation}
\mathcal{L}(\theta)
=
\mathcal{L}_{\mathrm{BC}}(\theta)
-
\alpha \,
\mathbb{E}_{s \sim \mathcal{D},\, a^\pi \sim \pi_\theta}
\big[
Q_\phi(s, a^\pi)
\big].
\label{eq:bfq_actor_loss}
\end{equation}

To normalize across dataset-specific Q-value scales, the coefficient $\alpha$ is further adapted as:
\begin{equation}
\alpha
=
\frac{\eta}{
\mathbb{E}_{(s,a) \sim \mathcal{D}}
\big[
\lvert Q_\phi(s,a) \rvert
\big]},
\end{equation}
where $\eta$ is a tunable hyperparameter. The normalization term is treated as a constant with respect to gradient updates.




\textbf{Q-Guided Action Selection at Test Time.}
To further improve test-time performance, we exploit the complementarity between the learned policy and value function through Q-guided action selection. Given a state $s$, we sample $K$ candidate actions in parallel from the learned policy,
$
a_1, \ldots, a_K \sim \pi_\theta(\cdot \mid s),
$
evaluate them with the learned Q-function, and select the highest-valued action:
\begin{equation}
\pi_\theta^{K}(s)
\;\triangleq\;
\arg\max_{a \in \{a_1,\ldots,a_K\}}
Q(s,a).
\end{equation}
Because both action sampling and Q-value evaluation are fully batched, this procedure adds only modest computational overhead in practice. As a result, Q-guided selection improves inference-time action quality without modifying training.
\begin{table*}[t]
\scriptsize
\centering

\setlength{\tabcolsep}{5.1pt} 
\resizebox{1.0\textwidth}{!}{
\begin{tabular}{l|ccc|ccc|cc}
\toprule
 & \multicolumn{3}{c}{\textbf{Non-Diffusion Policies}} & \multicolumn{3}{c}{\textbf{Multi-step Diffusion Policies}} & \multicolumn{2}{c}{\textbf{One-step Flow Policies}} \\
\midrule

Dataset  & BC & TD3-BC & IQL  & EDP & IDQL & DQL  & FQL & InsQL (Ours) \\

\midrule

 HalfCheetah-Medium-Expert  & 55.2 & 90.7 & 86.7 & 95.8 ± 0.1 & 91.3 ± 0.6 & 95.5 ± 0.1  & \textbf{99.8 ± 0.1} & 97.5 ± 0.1 \\
 Hopper-Medium-Expert       & 52.5 & 98.0 & 91.5   & 110.8 ± 0.4 & 110.1 ± 0.7 & \textbf{111.1} ± 0.4  & 86.2 ± 1.3 & 109.5 ± 1.3 \\
 Walker2d-Medium-Expert     & 107.5 & 110.1 & 109.6  & 110.4 ± 0.0 & 110.6 ± 0.0 & 111.6 ± 0.9  & 100.5 ± 0.1 & \textbf{112.9 ± 0.1} \\
\midrule

 HalfCheetah-Medium  & 42.6 & 48.3 & 47.4  & 50.8 ± 0.0 & 51.5 ± 0.1 & 52.3 ± 0.2  &  60.1 ± 0.1 &  \textbf{65.0} ± 0.1 \\
 Hopper-Medium       & 52.9 & 59.3 & 66.3 & 72.6 ± 0.2 & 70.1 ± 2.0 & 96.5 ± 1.3  & 74.5 ± 0.2 &   \textbf{100.5} ± 1.1 \\
 Walker2d-Medium     & 75.3 & 83.7 & 78.3  & 86.5 ± 0.2 & 86.1 ± 0.4 & 86.8 ± 0.0  &  72.7 ± 0.8 & \textbf{93.9} ± 0.1 \\
\midrule

HalfCheetah-Medium-Replay  & 36.6 & 44.6 & 44.2  & 44.9 ± 0.4 & 46.5 ± 0.3 & 47.9 ± 0.0  &   51.1 ± 0.1 & \textbf{52.1} ± 0.1 \\
Hopper-Medium-Replay       & 18.1 & 60.9 & 94.7 & 83.0 ± 1.7 & 99.4 ± 0.1 & 91.2 ± 0.0  &  85.4 ± 0.5 &  \textbf{102.1} ± 0.0 \\
 Walker2d-Medium-Replay     & 26.0 & 81.8 & 73.9  & 87.0 ± 2.6 & 89.1 ± 2.4 & \textbf{98.2} ± 0.1  &  82.1 ± 1.2 & 95.2 ± 0.6 \\
\midrule

Average (MuJoCo) & 51.9 & 75.3 & 77.0 & 82.4 & 84.1  &   89.0  &  79.2  & \textbf{92.1} \\
\midrule

AntMaze-Medium-Play    & 0.0 & 10.6 & 71.2   & 73.3 ± 6.2 & 67.3 ± 5.7 & 86.0 ± 1.8  &  78.0 ± 7.0 &  \textbf{90.0} ± 4.2 \\
 AntMaze-Large-Play   & 0.0 & 0.2 & 39.6  &  33.3 ± 1.9 & 48.7 ± 4.7 & 83.3 ± 2.5  &  84.0 ± 7.0 &  \textbf{89.1} ± 4.5 \\
AntMaze-Medium-Diverse   & 0.8 & 3.0 & 70.0  & 52.7 ± 1.9 & 83.3 ± 5.0 & \textbf{94.7} ± 2.5  &  71.0 ± 13.0 & 81.5 ± 4.1 \\
AntMaze-Large-Diverse   & 0.0 & 0.0 & 47.5  & 41.3 ± 3.4 & 40.0 ± 11.4 & 61.3 ± 8.4  &  \textbf{83.0} ± 4.0 & 80.2 ± 5.6 \\
\midrule

Average (AntMaze) & 0.2 & 3.5 & 57.1  & 50.2 & 59.8  &  81.3 &  79.0  & \textbf{85.2} \\

\bottomrule
\end{tabular}
}
\caption{Normalized performance on D4RL benchmarks spanning MuJoCo locomotion and AntMaze navigation. Boldfaced entries denote the best result in each row.}
\vspace{-10pt}
\label{tab:d4rl-results}
\end{table*}

\begin{figure}[htp]
    \centering
    \vspace{-0.3cm}
    \includegraphics[width=\linewidth]{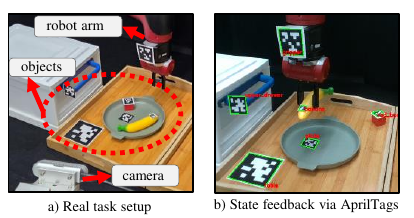}
    \caption{Real-World Robotic Manipulation Setup}
    \label{fig:workspace}
    \vspace{-0.5cm}
\end{figure}

\section{Experimental setting}

\subsection{Offline RL Benchmark.}
\begin{figure*}[htp]
    \centering    \includegraphics[width=\linewidth]{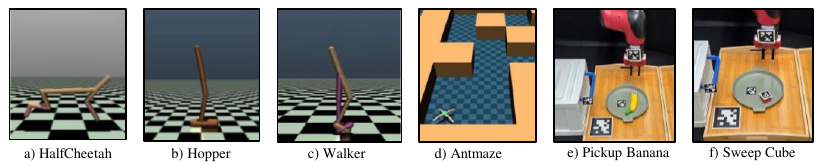}
    \caption{Evaluation tasks. (a--c) MuJoCo locomotion tasks (d) AntMaze Navigation task. (e--f) Real-world robotic manipulation tasks}
    \vspace{-0.5cm}
    \label{fig:mujocotask}
\end{figure*}

We evaluate InsQL on a diverse set of tasks from the D4RL benchmark suite \cite{fu2020d4rl}, a standard testbed for offline RL. Our experiments cover multiple domains, including locomotion and navigation, to assess generalization across different control regimes. MuJoCo locomotion tasks evaluate fine-grained continuous control under complex physical dynamics. Agents must maintain stability and efficiency while performing short-horizon decision-making in high-dimensional action spaces. AntMaze combines continuous control with long-horizon planning in maze-like environments. Agents must navigate to goal locations under sparse rewards, requiring coordination between low-level locomotion and high-level planning. Fig.~\ref{fig:mujocotask} (a-d) illustrates representative tasks used in our experiments.

\paragraph{Baselines}
We evaluate InsQL against a diverse set of competitive baselines spanning multiple policy classes. 
These include: 
(1) \emph{Non-diffusion methods}, namely Behavior Cloning (BC), TD3-BC \cite{fujimoto2021minimalist}, and IQL \cite{kostrikov2021offline}; 
(2) \emph{Multi-step diffusion-based approaches}, including IDQL \cite{hansen2023idql}, DQL \cite{wang2023diffusion}, and EDP \cite{kang2023efficient}; and 
(3) \emph{Single-step flow-based policies}, represented by FQL \cite{park2025flow}.

\subsection{Real-World Robotic Manipulation.}
We further evaluate InsQL on real-robot manipulation using a 7-DOF Sawyer arm with a parallel-jaw gripper in a structured tabletop workspace (Fig.~\ref{fig:workspace}). The scene contains both task-relevant and distractor entities (e.g., a banana, a cube, a plate/tray, and a drawer), each instrumented with AprilTags~\cite{olson2011apriltag}. A static RGB camera tracks tag poses, and all positions are expressed in the coordinate frame of a fixed table tag, yielding a low-dimensional, fully observed Markov state without loss of generality. Specifically, the state is a 10-D vector $s=[p_a,\,p_o,\,p_d,\,g_s]\in\mathbb{R}^{10}$ comprising the end-effector tag position $p_a\in\mathbb{R}^3$, the target object position $p_o\in\mathbb{R}^3$, a distractor pose $p_d\in\mathbb{R}^3$, and the gripper status $g_s\in\{0,1\}$. Actions are 5-D relative commands $a=[v_x,v_y,v_z,\omega_\psi,g_{\mathrm{cmd}}]\in\mathbb{R}^5$ (Cartesian velocity, yaw rate, and gripper command), where $g_{\mathrm{cmd}}$ is thresholded for binary actuation. We consider two manipulation tasks requiring coordinated reaching and grasping as illustrated in Fig.~\ref{fig:mujocotask} (e-f): \emph{Pickup Banana} (reach, grasp, and lift a toy banana to a target height) and \emph{Sweep Cube} (push a cube to a goal region on the tray). Success is defined by task-specific geometric thresholds on the tracked target pose (e.g., height for pickup and planar displacement for sweeping).

\paragraph{Dataset collection.}
For each task, we collect an offline dataset of 100 trajectories by combining 50 successful human teleoperated demonstrations with 50 autonomously generated suboptimal rollouts. Suboptimal data are mined either from a noisy BC policy trained on the human demonstrations or from a noisy random reaching PD controller, producing a multimodal mixture of near-success and failure behaviors. This combination provides both achievable solutions and informative failure modes, yielding a feasible and multimodal dataset.

\paragraph{Baselines.}
We compare InsQL against Behavior Cloning (BC) and DQL, reporting real-robot success rates over randomized initializations.

\subsection{Implementation Details.}
We implement InsQL on top of CleanDiffuser \cite{dong2024cleandiffuser}, following its evaluation protocol to ensure fair comparison with diffusion-based baselines implemented within the same framework.
The policy is parameterized as a 4-layer MLP with 256 hidden units per layer and Mish activations. The input consists of the concatenation of the action latent, state vector, and sinusoidal embeddings (dimension 64) of timesteps $t$ and $n$. The network outputs the average velocity. The critic employs two Q-networks, each implemented as a 4-layer MLP with 256 hidden units and Mish activations, taking concatenated state–action pairs as input and producing scalar Q-values.
Timesteps $t$, $n$, and $m$ are sampled uniformly subject to $0 \le n < m < t \le 1$. The primary hyperparameters are the \emph{conditioning ratio} $\lambda$ and coefficient $\eta$. Unless stated otherwise, we fix $\lambda = 0.5$ and tune $\eta$ over $\{0.001, 0.01, 0.05, 0.1, 0.3, 0.5, 1\}$. We use Adam with a learning rate of $3\times10^{-4}$. 

Results are reported as the average D4RL normalized score \cite{fu2020d4rl} over three seeds, with each trained policy evaluated on 150 episodes per task. All remaining hyperparameters follow the default DQL configuration in CleanDiffuser.

\section{Experimental result}
\begin{table}[tb]
\centering
\small
\begin{tabular}{l|c|c|c}
\toprule
\textbf{Method} & \textbf{Steps} & \textbf{Frequency} & \textbf{Training Time} \\
\midrule
InsQL (-) & 1  & 840.5 & \textbf{7.8} \\
InsQL  & 1  & 680.2 & \textbf{7.8} \\
\midrule
FQL & 1  & \textbf{929.6}    & 8.5 \\
\midrule
\multirow{4}{*}{DQL}
    & 5  & 238.1 & 11.7 \\
    & 10 & 150.1 & 16.1 \\
    & 20 & 75.2  & 24.8 \\
    & 50 & 35.5  & 49.5 \\
\bottomrule
\end{tabular}
\label{fig:training_inference_comparison}
\caption{Action frequency (Hz) and training wall-clock time (hours) for $10^6$ training steps, averaged over MuJoCo tasks, across methods and sampling steps. \textbf{InsQL (-)} denotes InsQL without Q-guided sampling.}
\vspace{-0.5cm}
\end{table}
The results on standard offline RL benchmarks are summarized in Table~\ref{tab:d4rl-results}, where we compare the proposed method with representative prior approaches.

\textbf{Locomotion (MuJoCo).} InsQL achieves strong and consistent improvements across locomotion tasks, outperforming competitive diffusion-based baselines such as DQL. On average, InsQL attains the highest score of 92.1, improving upon DQL from 89.0 to 92.1. The gains are particularly pronounced in the medium and medium-replay settings, where noisy and suboptimal datasets induce complex multi-modal action distributions that challenge conventional policy learning. These results underscore the importance of expressive policy modeling and stable value estimation.

Compared to acceleration-oriented variants such as EDP and IDQL, which improve efficiency at the expense of final performance, InsQL avoids this trade-off and delivers clear performance gains. Moreover, InsQL substantially outperforms the single-step FQL baseline (+12.9), demonstrating that compositional average-velocity modeling is more effective than distillation-based single-step policies.

\textbf{AntMaze.}
AntMaze tasks are particularly challenging due to sparse rewards and suboptimal demonstrations, making accurate Q-learning essential. As expected, BC underperforms, while Q-learning–based methods (IDQL, DQL) achieve stronger results. Among these, InsQL improves upon DQL from 81.3 to 85.2, outperforming all diffusion-based baselines and FQL.

Notably, although IDQL eliminates BPTT, it suffers a significant performance drop (81.3 to 59.8). In contrast, InsQL avoids BPTT within an actor–critic formulation while improving performance, demonstrating that compositional average-velocity modeling enhances both efficiency and effectiveness.

\textbf{Real-robot results.}
Table~\ref{tab:success_rate} reports success rates on two real-world manipulation tasks. Behavior Cloning (BC) fails on both tasks, highlighting the difficulty of direct imitation under noisy multimodal suboptimal demonstrations. In contrast, value-based generative policies substantially improve robustness: DQL attains $0.6$ on Sweep Cube and $0.9$ on Pickup Banana. InsQL further improves performance, achieving $0.7$ and $1.0$ success, respectively. We attribute these gains to InsQL’s one-step generative policy, which retains the expressiveness required for multimodal behavior while eliminating iterative denoising, and to its actor--critic optimization, which avoids BPTT and improves robustness under real-world noise and distribution mismatch.
\begin{table}[t]
\centering
\begin{tabular}{l|cc}
\toprule
\textbf{Method} & \textbf{Sweep Cube} & \textbf{Pickup Banana} \\
\midrule
BC    & 0.0 & 0.0 \\
DQL   & 0.6 & 0.9 \\
\midrule
InsQL & \textbf{0.7} & \textbf{1.0} \\
\bottomrule
\end{tabular}
\caption{\textbf{Real-robot success rates.} Success rate on Sweep Cube and Pickup Banana for BC, DQL, and InsQL (higher is better).}
\vspace{-0.5cm}
\label{tab:success_rate}
\end{table}
\section{Ablation Study}
\textbf{Training and Inference Efficiency.}
We compare wall-clock training time (1M steps) and decision frequency \cite{lu2025habitizing}, measured on a server equipped with an Intel Xeon Gold CPU (48 cores) and a single NVIDIA A100 (80GB) GPU. Training time is averaged across nine tasks. Decision frequency is computed over 3{,}000 action batches (batch size 2{,}500).

DQL exhibits near-linear growth in training cost as the number of denoising steps increases, rising from 11.7 hours (5 steps) to 49.5 hours (50 steps). In contrast, InsQL completes training in 7.8 hours. At inference, InsQL achieves 840.5 Hz without value-guided selection and 680.2 Hz with value guidance, showing only a minor reduction while substantially exceeding 5-step DQL (238.7 Hz) and 50-step DQL (35.5 Hz).

Compared to the single-step FQL baseline, InsQL attains comparable inference speed but requires less training time, as FQL relies on multiple NFEs to generate distillation targets. Despite similar efficiency, FQL consistently underperforms InsQL in policy performance. Overall, InsQL achieves both faster training and higher decision frequency without sacrificing expressiveness.



\begin{figure}[tpb]
    \centering 
    \includegraphics[width=\linewidth]  {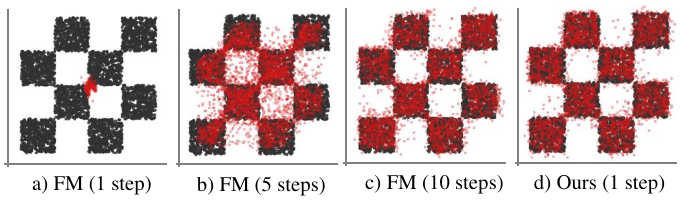}
    \caption{FM vs. InsQL on a multimodal checkerboard. FM requires multi-step generation to cover the modes and collapses under few-step evaluation, whereas InsQL achieves near-complete mode coverage and closely matches the target distribution in a single step.}
    \vspace{-0.6cm}
    \label{fig:fm_v_our}
\end{figure}

\textbf{Effect of the Conditioning Ratio.}
We analyze the influence of the boundary conditioning ratio $\lambda$ on HalfCheetah performance (Table~\ref{tab:flowrate}). The optimal results are obtained at $\lambda = 0.5$, achieving scores of 97.5 (Medium-Expert), 65.0 (Medium), and 52.1 (Medium-Replay). In contrast, extreme settings ($\lambda=0$ or $\lambda=1$) significantly degrade performance, with $\lambda=0$ (no boundary conditioning) failing to produce meaningful policy learning. These findings indicate that moderate boundary conditioning serves as an effective regularization mechanism, stabilizing training and improving robustness.
\begin{table}[htb]
\centering
\vspace{-0.3cm}
    \small
    \begin{tabular}{@{}c|ccccc@{}}
    \toprule
    $\lambda$     & \multicolumn{1}{r}{1} & \multicolumn{1}{r}{0.75} & \multicolumn{1}{r}{0.5} & \multicolumn{1}{r}{0.25} & \multicolumn{1}{r}{0} \\ \midrule
    Medium Expert & 79.8                  & 95.5              & \textbf{97.5}           & 93.1                    & 1.5                \\
    Medium        & 43.3                  & 63.8               & \textbf{65.0}           & 61.5                    & -1.5                  \\
    Medium Replay & 48.0                  & 49.7                     & \textbf{52.1}           & 51.3                       & 9.6                  \\ \bottomrule
    \end{tabular}
    \caption{Ablation study on the boundary conditioning ratio $\lambda$ for HalfCheetah tasks under different dataset regimes. Results are reported as D4RL normalized scores.}
    \label{tab:flowrate}
    \vspace{-0.3cm}
\end{table}

\textbf{Expressiveness of FM and InsQL Policies.}
Prior work on diffusion-based Q-learning (DQL) indicates that policy expressiveness is critical for strong actor--critic performance. To assess expressiveness under single-step generation, we compare FM and InsQL on a synthetic checkerboard dataset with a highly multimodal distribution. As shown in Fig.~\ref{fig:fm_v_our}, FM requires multiple generation steps to achieve adequate coverage and suffers from mode collapse under few-step evaluation. In contrast, InsQL achieves accurate mode coverage and closely matches the target distribution in a single step.

\textbf{Effect of Q-guided sampling.}
\begin{figure}[t]
    \centering
\includegraphics[width=\linewidth]{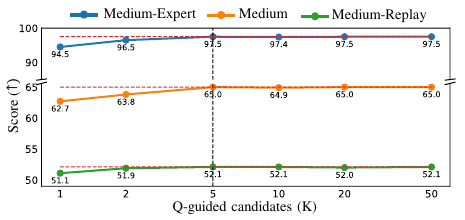}
    \caption{Ablation on the effect of Q-guided sampling on HalfCheetah. $K$ denotes the number of candidate actions sampled per state; we choose the action with the highest predicted $Q$.}
    \label{fig:ablation_q_guided_k}
    \vspace{-0.8cm}
\end{figure}
We study how the number of Q-guided candidates $K$ affects performance.
Fig.~\ref{fig:ablation_q_guided_k} reports results on the three HalfCheetah settings.
Increasing $K$ from $1$ to $5$ yields consistent gains across all datasets (e.g., $\,+3.0$ on Medium-Expert and $\,+2.3$ on Medium), indicating that a small amount of candidate refinement already provides substantial benefit.
Beyond $K=5$, performance largely saturates, suggesting diminishing returns from evaluating additional candidates.
Overall, we set $K=5$ as a strong default that balances performance and computation.

\section{Conclusion}

In this paper, we introduced \textbf{Instant Flow Q-Learning (InsQL)}, an end-to-end offline RL framework that enables \emph{single-step} action generation by modeling compositional \emph{average velocity} rather than marginal velocity fields. This design eliminates solver unrolling and BPTT while retaining high expressiveness, and can be further improved at test time via lightweight Q-guided candidate selection. InsQL achieves state-of-the-art or competitive results on D4RL MuJoCo and AntMaze with substantially lower training cost and higher decision frequency, and also performs strongly on real-robot tabletop manipulation. Overall, InsQL demonstrates that single-step flow-based policies offer a practical and scalable alternative to multi-step diffusion policies for offline and resource-constrained settings.

\textbf{Limitation.} This work focuses on state-based Offline RL tasks, which avoid confounding factors such as visual feature extraction, multimodal fusion, and architecture-specific biases, allowing a cleaner evaluation of InsQL’s algorithmic contributions. Extending InsQL to more challenging settings, including offline-to-online reinforcement learning, visual observation-based control, and broader applications
\cite{vu2022softgroup,vu2023scalable,nguyen2023dimcl},
remains an important direction for future work.


\bibliography{main}
\bibliographystyle{IEEEtran}
\end{document}